\documentclass[runningheads]{llncs}

\usepackage[T1]{fontenc}
\usepackage{graphicx}
\usepackage{color, wrapfig}
\usepackage{tabularx}
\usepackage{multirow}
\usepackage{caption}
\usepackage{subcaption}
\usepackage{amsmath}
\usepackage{float}
\usepackage{marvosym}

\begin{document}
\title{Reproducibility in Machine Learning-Driven Research}
\titlerunning{Reproducibility in Machine Learning-Driven Research}

\author{
Harald Semmelrock\inst{2} \and
Simone Kopeinik\inst{1} \and
Dieter Theiler\inst{1} \and
Tony Ross-Hellauer\inst{1,2} \and
Dominik Kowald$^{\textrm{(\Letter)}}$\inst{1,2}}

\institute{Know-Center GmbH, Graz, Austria \\\email{{skopeinik,dtheiler,tross,dkowald}@know-center.at} \and
Graz University of Technology, Graz, Austria \\\email{harald.semmelrock@student.tugraz.at}
}

\authorrunning{Semmelrock, Kopeinik, Theiler, Ross-Hellauer, and Kowald}

\maketitle

\begin{abstract}
Research is facing a reproducibility crisis, in which the results and findings of many studies are difficult or even impossible to reproduce. 
This is also the case in machine learning (ML) and artificial intelligence (AI) research. Often, this is the case due to unpublished data and/or source-code, and due to sensitivity to ML training conditions. Although different solutions to address this issue are discussed in the research community such as using ML platforms, the level of reproducibility in ML-driven research is not increasing substantially. 
Therefore, in this mini survey, we review the literature on reproducibility in ML-driven research with three main aims: (i) reflect on the current situation of ML reproducibility in various research fields, (ii) identify reproducibility issues and barriers that exist in these research fields applying ML, and (iii) identify potential drivers such as tools, practices, and interventions that support ML reproducibility. 
With this, we hope to contribute to decisions on the viability of different solutions for supporting ML reproducibility.

\end{abstract}

\section{Introduction}
\label{chap:introduction}

Similar to other scientific fields~\cite{baker_1500_2016}~\cite{sayre_reproducibility_2018}, research in artificial intelligence (AI) in general, and machine learning (ML) in particular, is facing a reproducibility crisis~\cite{hutson_artificial_2018}.
Here, especially unpublished source-code and sensitivity to ML training conditions make it nearly impossible to reproduce existing ML publications, which also makes it very hard to verify the claims and findings stated in the publications.\\\\
One potential solution for enhancing reproducibility in ML is the use of ML platforms such as OpenML, Google Cloud ML, Microsoft Azure ML or Kaggle. 
However, in a recent study~\cite{gundersen_machine_2022} found that the same experiment executed on different platforms leads to different results.
This suggests that still a lot of research is needed until out-of-the-box reproducibility can be provided.
However, a systematic overview of the literature on ML reproducibility is still missing, especially with respect to the barriers and drivers of reproducibility that can be found in the literature.
An example of a driver could be code sharing or hosting reproducibility tracks/challenges at scientific conferences~\cite{gibney_this_2019}. One example for this is the reproducibility track at the European Conference on Information Retrieval (ECIR)~\cite{kowald2020unfairness,muellner2021robustness}\\
With respect to potential barriers, it is still not clear to what extent the use of ML could even fuel reproducibility issues~\cite{gibney_could_2022}, e.g., via bad ML practices such as data leakage~\cite{kapoor_leakage_2022}.\\\\
This work aims to provide an overview of the situation and identify the different drivers and barriers present. This should allow for a  better understanding of the following three aspects:
\begin{itemize}
    \item The situation of ML reproducibility in different research fields (see Section~\ref{chap:situationsubfields}). 
    \item Reproducibility issues that exist in research fields applying ML, and the barriers that cause these issues (see Section~\ref{reproducibilityissues}).
    \item The drivers that support ML reproducibility, including different tools, practices, and interventions (see Section~\ref{chap:drivers}).
\end{itemize}


\subsection{Degrees of Reproducibility}
According to~\cite{gundersen_state_2018}, there are three degrees of reproducibility in ML, which can be seen in Table \ref{tbl:different-types-reproducibility}.

\begin{table}[H]
	\caption[Different degrees of ML reproducibility according to~\cite{gundersen_state_2018}]{\textbf{Different degrees of reproducibility according to~\cite{gundersen_state_2018}} } 
	\label{tbl:different-types-reproducibility}
	\begin{center}
	\begin{tabular}{lp{7.5cm}}
	\textbf{Type} & \textbf{Requirement} \\
	(R1) Experiment Reproducibility & The same implementation (including same software versions, hyperparameters, etc.) of the ML
method (i.e., the algorithm) must produce (exactly) the same results when using the same training and test
data. If one produces different results, then only the hardware where the ML experiment was
reproduced could be the reason for this difference.\\
	(R2) Data Reproducibility & An alternative implementation of the ML method must produce (almost) the same results when executed using the same
data. If there are differences in the results, then probably differences in the concrete implementation
(e.g., different versions of a library) are the reasons for this.\\
(R3) Method Reproducibility & An alternative
implementation of the ML method executed on different data must produce the same results (or at least
findings).\\
	\end{tabular}
	\end{center}
\end{table}

\begin{figure}[H]
	\begin{center}
		\includegraphics[width=.8\textwidth]{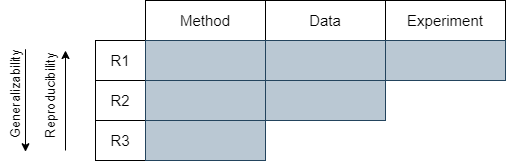}
	\end{center}
	\caption[Degrees of reproducibility]{\textbf{Degrees of reproducibility}. Adapted from~\cite{gundersen_state_2018}}
	\label{fig:degrees}
\end{figure}

We see that R1 is concerned with the exact reproduction of results, i.e., output, over multiple runs of the same ML method implementation and data. This is also often referred to as computational reproducibility. R2, however, is a bit more general, in a way that the ML method is implemented differently, but should still be able to produce almost the same results given the same data. This ``Data Reproducibility'' requirement makes sure that the information drawn from the data is consistent and not too heavily dependent on minor implementation differences.
R3 generalizes this even more and is only concerned with the findings. Relying only on the use of the same ML method, findings should always be consistent, no matter the exact implementation or exact data. 
Accordingly, R3 leads to the highest form of generalizability but also to the weakest form of reproducibility. This is conversely true for R1, which leads to the highest form of reproducibility and lowest form of generalizability.
This interplay between generalizability and reproducibility of the different degrees can be seen in Figure~\ref{fig:degrees}.

Furthermore, the structure of Figure \ref{fig:degrees} gives information about what building blocks the different degrees of reproducibility are concerned with. Thus, R3 is only concerned with the ML method, R2 is concerned with the ML method and data and R1 requires all three building blocks: the ML method, the data, and the experiment. This is where the names of the degrees come from.


\subsection{Reproducibility versus Replicability}

The two terms Reproducibility and Replicability are often used interchangeably in science, even though the difference is often crucial.\\
For computer science and ML, the Association for Computing Machinery (ACM) distinguishes the terms in the following way:
\begin{itemize}
    \item Reproducibility - the results can be obtained by a different team with the same experimental setup
    \item Replicability - the results can be obtained by a different team with a different experimental setup
\end{itemize}

These definitions harmonize with the definitions seen in the literature~\cite{beam_challenges_2020}. Comparing the definitions to the different degrees of reproducibility coined by~\cite{gundersen_state_2018}, we can see that degree R1 is somewhat similar to what is here referred to as Reproducibility, and both R2 and R3 can be seen as similar to what is referred to as Replicability.
For the purpose of this work, the definitions concerning the three degrees of reproducibility~\cite{gundersen_state_2018} will be used.


\section{ML Reproducibility in Different Research Fields}
\label{chap:situationsubfields}
We mainly focus on two research fields, in which ML reproducibility is discussed: (i) Computer Science, and (ii) health / life science. Additionally, we mention a few other research fields, which have seen similar trends in connection with the reproducibility crisis.

\subsection{Computer Science}
In Computer Science, reproducibility is mainly discussed in two specific sub-fields of ML i.e., Deep learning and Reinforcement learning, and in two application fields of ML i.e., Natural language processing and Recommender systems. 

\subsubsection{Deep learning}
Neural networks are widely used in ML and can be applied to both supervised and unsupervised learning tasks.
Neural networks, however, are known to be inherently non-deterministic and produce different results on multiple reruns, due to the many sources of randomness during training~\cite{ahmed_managing_2022}. Not being able to get the same results using the exact same code and data is a big challenge to reproducibility, specifically w.r.t. degree R1. These different outcomes, however, may not necessarily be statistically different, such that the same conclusions can still be drawn. This is a distinction, researchers in the area of using ML for imaging have come to~\cite{piantadosi_reproducibility_2019}. Their research showed that while the performance of Convolutional Neural Networks (CNN) on image segmentation shows slightly different results on multiple reruns, the obtained results are not statistically different, and the findings are robust.\\
In order to counteract the effect of different outputs on multiple reruns in general, the idea of controlling the sources of randomness has been proposed~\cite{ahmed_managing_2022}.\\
To add to this,~\cite{shahriari_how_2022} found that the concrete versions of deep learning frameworks, such as PyTorch or TensorFlow, can have a big impact on performance, in a way where upgrading the version of a model being used can drastically increase or decrease the performance.

\subsubsection{Reinforcement learning}
Reinforcement Learning (RL) is the subfield of ML, where reproducibility issues are seemingly discussed the most.
RL is especially susceptible to reproducibility issues, partially because of additional sources of non-determinism in the learning process, which other areas are not subjected to~\cite{nagarajan_impact_2018}.
To overcome this obstacle, frameworks, which provide reproducible environments, are commonly used in reinforcement learning research~\cite{brockman_openai_2016}~\cite{ferigo_gym-ignition_2020}~\cite{young_minatar_2019}. These frameworks act as testbeds where different RL algorithms can be evaluated in a common environment.
Additionally, different papers have outlined the implementation of deterministic reinforcement learning algorithms, by controlling all the sources of non-determinism~\cite{nagarajan_deterministic_2019}~\cite{abel_simple_nodate}.\\
Furthermore,~\cite{khetarpal_re-evaluate_2018} propose to standardize an evaluation pipeline, which can be used for reproducible benchmarking of different reinforcement algorithms, similar to competitions like Pommerman and the Learning to Run Challenge.

\subsubsection{Natural language processing}
A review of the status quo in terms of reproducibility in Natural language processing (NLP) by~\cite{belz_systematic_2021} highlights some core issues, which overlap with the ones mentioned for neural networks. Even simple rerunning of code can yield different results, again undermining reproducibility of degree R1.\\ 
When trying to reproduce NLP research, it was evident that even minor changes to parameter settings, such as rare-word thresholds, can drastically change performance. One striking finding, among others, was that whenever the results of a reproduction deviated from the original reported results, the reproduced version performed worse than the original.\\
A different paper by~\cite{dror_replicability_2017} tried to tackle these performance benchmarking issues and proposed a statistically sound replicability analysis framework, where NLP algorithms are benchmarked against each other using multiple different datasets. This is done in a way that not a specific dataset is picked for evaluation, on which the algorithm performs well, but multiple datasets are used to measure performance.

\subsubsection{Recommender systems}
Recommender systems research has also been affected by the reproducibility crisis. The research area has observed that a stagnation in progress could happen due to the difficulty of benchmarking new solutions against existing solutions. This has created a form of ``phantom progress'' where it is not evident whether proposed new methods actually perform better than traditional ones~\cite{cremonesi_progress_2021}. 
A study by~\cite{beel_towards_2016} has evaluated the reasons for this and proposed solutions to the problem, which heavily overlap with the proposals of other research areas, e.g., modern publication practices that foster the use of reproducibility frameworks, and the establishment of best-practice guidelines.


\subsection{Health and Life Science}
There is a debate that the widespread adoption and use of ML has fueled reproducibility issues in health and life science~\cite{gibney_could_2022}. Accordingly, domain-specific research has been done to address and handle these reproducibility problems. 
While health and life science may have specific problems and solutions for reproducibility, a review of the state-of-the-art is important to come to a conclusion about reproducibility in ML as a whole.\\\\
Areas of health / life science making use of ML (often coined ML4H) have had a lot of trouble with missing data when trying to reproduce a research result. Data in the medical area is often private and cannot be publicized nor shared~\cite{beam_challenges_2020}~\cite{mcdermott_reproducibility_2019}.\\
Importantly, however, health and medical science are fields where reproducibility is of critical importance, since the verification of results is important before ML results can be used clinically~\cite{beam_challenges_2020}. 
To tackle the problem and overcome the reproducibility crisis in health / life science, a solution has been proposed by~\cite{mcdermott_reproducibility_2019}. This solution addresses three main stakeholders: the data providers, the journals / conferences and the ML4H research community. The idea is that the three stakeholders should  interact in rigorous and open collaboration.
This solution, however, relies on the collaborative sharing of data, which still has privacy issues connected to it. Possible solutions to this have also been mentioned by~\cite{mcdermott_reproducibility_2019}, such as (i) Privacy-Preserving Analysis techniques or specific data collection regimes, e.g., the Verily Project Baseline.\\
In general, this proposal follows the ideas of the FAIR data principles (findability, accessibility, interoperability, and reusability) and appeals that the scientific data of the area should adhere to these FAIR principles~\cite{colliot_reproducibility_2022}.\\\\
In addition to this, many domain-specific tools, such as NiLearn, Clinica or OSF have been proposed to help with making the research more reproducible~\cite{colliot_reproducibility_2022}.

\subsection{Other Research Fields}
Apart from computer science and health / life science, many different research fields have benefitted tremendously from ML, but, however, have recently also faced issues regarding reproducibility. Most of these reproducibility issues stem either from data leakage~\cite{barnett_genomic_2022} or the lack of computational reproducibility. Across different research fields applying ML, the need for reproducibility is discussed in literature, a few of which are highlighted below:

\subsubsection{Chemistry}
ML is used very commonly to interpret patterns in data for chemistry. To keep reproducibility issues at a minimum, best practices have been proposed by~\cite{artrith_best_2021}, which introduce guidelines to follow and inform about best practices, concerning, e.g., code and data sharing, and data leakage.

\subsubsection{Materials science}
ML models are used for the informed design of new materials. Recently, however, these workflows have increasingly faced reproducibility issues, especially effects of randomness during training and lack of reproducibility across platforms and versions~\cite{pouchard_replicating_2020}.
    
\subsubsection{Genomics}
A systematic review by~\cite{barnett_genomic_2022} has found that reported results are often inflated in recent research and are hard to evaluate, since data leakage is a prevalent issue in the research field. Galaxy is a biomedical research platform, which is used in genomics and biomedical research for running analyses and aims to improve reproducibility~\cite{goecks_galaxy_2010}. 
    
\subsubsection{Satellite imaging}
All kinds of ML techniques, e.g.,  unsupervised learning techniques, supervised learning techniques and deep learning are used within image segmentation for satellite imaging. The Monte-Carlo cross-validation is often used to verify new methods, however has been found to be very susceptible to data leakage~\cite{nalepa_validating_2019}, which negatively impacts reproducibility. 

\section{Barriers in ML Reproducibility}
\label{reproducibilityissues}

When it comes to the different barriers associated with reproducibility in ML, it is important to make a distinction between the different degrees of reproducibility. Some barriers only affect certain degrees of reproducibility. Computational problems, as an example, which prevent researchers from reproducing the exact same results using the same code and data is only a concern for reproducibility of degree R1. On the contrary, degrees R2 and R3 are more concerned with the findings and conclusions than with the exact outputs.

\subsection{Computational Problems}
Recent studies have shown, that sharing of code and data alone is not enough to ensure reproducibility even of degree R1.
In some cases, not even the assistance of the original author was enough to reproduce the original results~\cite{hardwicke_data_2018}.
This barrier of computational reproducibility can be attributed to a few factors, which are discussed in the following.

\subsubsection{Inherent nondeterminism}
The inherent nondeterminism in ML is a major reason why reproducibility of degree R1 cannot be easily achieved.
Even if both the data and code used in an experiment are known, the pseudo-random numbers generated throughout the training of the ML model can heavily alter its results~\cite{ahmed_managing_2022}.
Because of this, fixed random number seeds can be vital to the reproducibility of ML. 

\subsubsection{Environment differences}
Studies have shown that both hardware differences, such as different GPUs or CPUs, as well as different compiler settings can result in different computation outcomes~\cite{hong_evaluation_2013}.
Furthermore, a comparison between the same ML algorithm with fixed random seeds that was executed using PyTorch and TensorFlow also results in different results~\cite{pouchard_replicating_2020}.\\
Additionally, a survey, which was conducted by~\cite{gundersen_machine_2022} to investigate whether different ML platforms, such as OpenML or Kaggle, provide out-of-the-box reproducibility, also uncovered reproducibility issues.

\subsection{Missing Data and Code}
A study by~\cite{hutson_missing_2018} found that published ML research is often not accompanied by available data and code. Only one third of researchers share the data, and even fewer share the source code. This can have a lot of reasons, such as private data or code that itself is based on unpublished code. Furthermore, the problem may also be attributed to the increasing pressure for researchers to publish quickly, which in turn does not allow them to polish the code and decreases the willingness to release the code.

\subsection{Methodological Problems}
When computational problems are handled and code and data are available, then the reproducibility of degree R1 of concrete experiments should, in theory, be possible.
However, in reality, the results are often based on methodological errors made  throughout the experiment. One well-known methodological error is data leakage, which can come in different forms~\cite{kapoor_leakage_2022}. 
Data leakage has become a widespread issue of research applying ML recently~\cite{kapoor_leakage_2022}.
This can be attributed to the increasing amount of non-experts using ML for various research fields~\cite{gibney_could_2022}, which is fueled by the ease of application of auto-ML libraries and no-code off-the-shelf AI tools.
In essence, data leakage happens when some form of data, which the ML model should not be trained on, leaks into the training process. Data leakage can be categorized into 3 subcategories.

\subsubsection{L1 - No clean train/test split}
The subcategory L1 summarizes the most obvious cases of data leakage and is further split into 4 variants: (1) the training data and test data are not split at all, (2) the test data is also used for feature selection, (3) the test data is also used for imputation during preprocessing, and (4) there are duplicates in the dataset, which occur in both the test and training data.

\subsubsection{L2 - Use of non-legitimate data}
Subcategory L2 is concerned with training data, which contains non-legitimate features. For example, when the use of anti-hypertensive drugs is used to predict hypertension~\cite{kapoor_leakage_2022}. This data is non-legitimate, since it would not be available in a real world scenario and cannot realistically be used to predict hypertension for a new patient. Features, which result in data leakage of subcategory L2 are often synonymous with the target variable. Generally, deciding whether a feature is non-legitimate for a specific task requires a lot of domain-knowledge.

\subsubsection{L3 - Test set is not drawn from distribution of scientific interest}
L3 consists of 3 different types of data leakage. Firstly, temporal leakage is concerned with ML models, which try to predict future outcomes. In that case, temporal leakage happens when some training samples have a later timestamp than samples available in the test set. This would mean that the ML model is being trained on information coming ``from the future''. Accordingly, the ML model makes use of information, which it cannot have in a realistic scenario, and therefore performs better than it would in reality.
Secondly, the train and test data need to be independent of each other. This means that there cannot be samples in both the train and test data which are, e.g, from the same person. Thirdly, the test set should not be drawn selectively. An example of this is when the model is only evaluated on specific data, which it performs well on, and then these results are generalized to all use cases. Doing this would result in a selection bias.

\subsection{Structural Problems}

\subsubsection{Privacy concerns}
Availability of data is key for the reproducibility of any supervised or unsupervised ML algorithm. Even for reproducibility of degree R3, where different data should yield similar conclusions. In this case, it is important to have some representative dataset, which holds comparable information.
However, the public release of data is not always an option, since the data can be private, such as health-related patient data or user data collected by companies.\\
To counteract this, privacy preserving technologies have been proposed~\cite{mcdermott_reproducibility_2019}. There are a few different ways how privacy preserving techniques could work, e.g., to allow the training of ML models without sharing the data, and to work with encrypted or simulated data.

\subsubsection{Competitive advantage}
Comparing the status-quo in reproducibility of ML in academic research versus industrial applications gives interesting insights~\cite{gundersen_standing_2019}.
While both academia and industry still have a lot of work to do in terms of supporting reproducibility, there are different main reasons why reproducibility is not ensured in these two areas.
Academic research may just lack enough incentive or rewards for the increased effort of making sure the results are reproducible.
In contrast to this, in industrial applications it is often a problem that providing reproducibility may come with a decreased competitive advantage since. 

% ...

\section{Drivers for ML Reproducibility}
\label{chap:drivers}

After review of the status-quo in the different research fields, combined with the identified barriers, we now present an outline of the main proposals / drivers to address reproducibility. These drivers could be of critical help in overcoming the reproducibility crisis.

\subsection{Standardized Environments}
Container software, such as e.g., Docker, could be used to standardize the ML environment and allow for simple sharing of the container, i.e., environment and code, with other researchers~\cite{boettiger_introduction_2015}.
Furthermore, Code Ocean is a computational research platform, which has been specifically developed and tailored towards the needs of researchers. Because of this, it allows researchers to focus on the research questions instead of the standardization of environments~\cite{clyburne-sherin_computational_2019}.

\subsection{Checklists and Guidelines}
The idea of checklists and guidelines has been applied effectively in the past, e.g., for safety-critical systems.
To build on top of that,~\cite{pineau_improving_2021} proposed a ML reproducibility checklist, which should ensure the inclusion of necessary information for reproducibility. This checklist has been found promising during the NeurIPS 2019 reproducibility challenge, and has been suggested as best practice by researchers of different fields~\cite{artrith_best_2021}. In Addition to the use of checklists to make sure specific requirements are met by different papers (such as proper scientific method or achievement of certain reproducibility standards etc.), \cite{liu_reviewergpt_2023} propose to use Large Language Models (LLMs) to verify these checklists and use them to review papers in general. By doing this, both the efficiency, and the accuracy of verifying checklists can be improved, thus further enhancing the viability of checklists. \cite{liu_reviewergpt_2023} found that GPT-4 is the LLM, which stood out amongst others like Bard, Alpaca or LLaMa and is especially capable of doing this task and was able to achieve an 86.6\% accuracy of verifying the fulfillment of different checklist questions.
\subsection{Model Info Sheets}
Similar to checklists, model info sheets are questionnaires specifically tailored towards handling data leakage, i.e., detection and prevention. 
\cite{kapoor_leakage_2022} propose these so-called model info sheets, which should be filled and published with the work, in order to allow other researchers to quickly verify the data used to train the ML model. To do this, the model info sheets require the authors to answer detailed questions about the data and respective train/test splits, which are aimed towards each type of data leakage\cite{kapoor_leakage_2022}.\\
Model info sheets seem like a promising solution for a lot of types of data leakage, given that research of non-experts in ML often falls prey to different types of data leakage~\cite{kapoor_leakage_2022}. Additionally, the model info sheets are a low-effort solution for the problem they are trying to solve.

\subsection{Awareness}
Awareness of the reproducibility crisis and informing about ways to support reproducibility can be a very strong factor in overcoming the crisis, as also shown by the reproducibility crisis in Psychology~\cite{wiggins_replication_2019}.\\
Many different initiatives have been started to increase awareness on the reproducibility crisis, a few of which are highlighted below:
\begin{itemize}
    \item The ReScience journal publishes peer-reviewed papers, which are specifically about trying to reproduce results of original publications. These reproductions are then made available on GitHub for other researchers~\cite{rougier_sustainable_2017}.
    \item ReproducedPapers.org is a Webs ite with a similar idea but focuses on teaching. It is an online repository for papers, which are reproducible. The ReproducedPapers initiative also aims to include the reproduction of at least one ML paper in the curriculum of ML students~\cite{yildiz_reproducedpapersorg_2021}.
    \item Reproducibility challenges, where several researchers try to reproduce many recent publications in parallel, are being held yearly. These challenges allow for an analysis of the success rate of reproduction and can be used to evaluate progress over multiple years~\cite{pineau_improving_2021}.
\end{itemize}

\subsection{Journals}
To increase the awareness and enforce a minimum of reproducibility standards, journals could adjust their requirements for publication.
Recently, it has already become a standard procedure in many journals to require data and/or code availability for publication~\cite{pineau_improving_2021}~\cite{peng_reproducibility_2015}~\cite{hardwicke_data_2018}.\\
Furthermore, to address the problem of researchers deliberately tweaking results for a higher likelihood of being published, journals could introduce the possibility of preregistration. Preregistration allows researchers to submit their research intention ahead of time to become registered for the publication in the future. The journal decides whether to publish the paper based on the research plan, and researchers then do not have to worry about the outcome of the experiments, which should improve the credibility of the findings~\cite{stromland_preregistration_2019}~\cite{nosek_preregistration_2019}.\\
The ACM TORS journal (Transactions on Recommender Systems), is a good example of promoting reproducibility in ML. Firstly, it allows for preregistration, and secondly, it also allows for publication of ``reproducibility papers'', which are specifically concerned with reproduction studies and tools for enhancing reproducibility.
\section{Conclusion and Future Work}
\label{chap:conclusions}

Almost all areas of ML, as well as research fields using ML (e.g., health and life science), have seen signs of a reproducibility crisis. 
Appropriately, this reproducibility crisis has received increasing attention recently and is being discussed in research. 
It is time to tackle this reproducibility crisis, which at first needs a common understanding of the definitions and different degrees of reproducibility.\\

This work gave an overview over the current state-of-the-art in different research areas.
Additionally, it gave an insight into the different barriers and drivers for ML reproducibility.
For the continuation of this work, we will dive deeper into the concepts outlined so far and will compare them across different research fields. 
With this, we hope that a decision on the viability of different solutions can be supported.  

\subsubsection{Acknowledgements.} This research is supported by the Horizon Europe project TIER2 under grant agreement No 101094817.




\bibliographystyle{splncs04}
\bibliography{ReproducibilityML_expose}

	
\end{document}